\documentclass[11pt,a4paper]{article}

\usepackage[T1]{fontenc}
\usepackage{lmodern}
\usepackage[margin=1in]{geometry}
\usepackage{microtype}
\usepackage{amsmath,amssymb}
\usepackage{booktabs,multirow,array}
\usepackage{graphicx}
\usepackage{xcolor}
\usepackage{caption}
\usepackage{subcaption}
\usepackage{enumitem}
\usepackage{float}
\usepackage{tikz}
\usetikzlibrary{positioning,arrows.meta,calc,decorations.pathreplacing}
\usepackage[numbers,sort&compress]{natbib}
\usepackage[colorlinks=true,linkcolor=blue!50!black,citecolor=blue!50!black,urlcolor=blue!50!black]{hyperref}

\setlist{itemsep=1pt,topsep=3pt}
\newcommand{\code}[1]{\texttt{#1}}

\title{\bfseries Crop Yield Prediction for Punjab, Pakistan:\\
A Tree-Ensemble and Leaf-Health Prototype, and What Random Validation Hides}

\author{
\parbox{\textwidth}{
\centering
Amina Asif$^{1,*}$ \quad
Qurat Ul Ain Asif$^{2}$ \quad
Noor Bakhat Asif$^{3}$ \quad
Kasa-e-Fatima$^{4}$\\[5pt]
{\small
$^{1}$ Department of Artificial Intelligence, The Islamia University of Bahawalpur, Pakistan\\
$^{2}$ Key Laboratory of Xinjiang Endemic Phytomedicine Resources, Ministry of Education,\\
School of Pharmacy, Shihezi University, Shihezi 832002, China\\
$^{3}$ Department of Physics, Government College of Women University of Faisalabad, Pakistan\\
$^{4}$ Department of Artificial Intelligence, The Islamia University of Bahawalpur, Pakistan\\[4pt]
$^{1}$ asifyaqoobamina@gmail.com\\
$^{2}$ quratulainasif7117@stu.shzu.edu.cn\\
$^{3}$ noorbakhatasif@gmail.com\\
$^{4}$ kasaefatima81@gmail.com
}
}
}

\date{}

\begin{document}
\raggedbottom
\maketitle

\begin{abstract}
Yield forecasts help planners and farmers decide on inputs, storage and imports, but small agricultural tables make it easy to report accuracy that does not survive a new season. We built a crop-yield prototype for Punjab, Pakistan that combines Random Forest, XGBoost, support vector regression and a Ridge-stacked ensemble with a MobileNetV2 leaf-health classifier, and deployed it as a web application with SHAP explanations. A random 80/20 split of the merged Kaggle-derived table (414 rows) gives the ensemble an $R^2$ of 0.991. An audit showed that this table holds only 46 independent observations: a join with nine temperature records per year had copied every crop-year nine times. Holding out whole years takes XGBoost on the same rows from $R^2 = 0.994$ to $-0.20$. On the deduplicated data, and again on a longer table that we rebuilt from FAOSTAT (1990--2024, 70 observations), a per-crop linear trend (leave-one-year-out RMSE 0.29 Ton/Ha on the FAOSTAT table) beats every model that is not given the year (0.84--1.06), and neither pesticide use nor national temperature change explains what the trend misses. An apparent gain from pesticide use in the Kaggle table disappeared on the FAOSTAT table, where the pesticide series turns out to be mostly imputed; the two releases of that series disagree, and both look like interpolation. An independent district-level wheat panel (36 districts, 13 seasons) shows that most variation in Punjab is spatial and that a district mean with a common trend matches the learned models. The leaf classifier reaches 99.87\% accuracy on held-out PlantVillage images and recalls 96.4\% of 336 unseen diseased leaves after near-duplicates of its training images were removed. However, no leaf image is paired with a yield record, so the health score used to train the yield models had to be constructed, and it adds nothing. We report these negative findings together with the prototype.
\end{abstract}

\section{Introduction}
Agriculture contributed about 24\% of Pakistan's GDP and 37.4\% of its employment in 2023--24 \citep{economicsurvey2024}, and Punjab is one of its main wheat- and maize-growing regions. A farmer deciding how much fertiliser to buy, or a planner deciding how much grain to import, would like an estimate of the coming yield. Machine learning is a natural tool for this, and the literature on it is large \citep{vanklompenburg2020}, with random forests \citep{jeong2016} and deep networks \citep{khaki2019} among the methods applied. Much of it, though, is evaluated on tables small enough that a careless split can make a weak model look excellent, a failure that is well documented in data mining generally \citep{kaufman2012}.

This paper started as a coursework project with a conventional goal: predict wheat and maize yield in Punjab from climate, pesticide and soil variables, compare several regressors, and wrap the best one in a web application. We later added a computer vision module that reads a leaf photograph and returns a health score, with the idea that visible disease should say something about yield. The first version of our results looked very good. A stacking ensemble scored $R^2 = 0.991$ on a held-out test set. We were pleased, and then we looked at the data more carefully.

What we found changes how those numbers should be read, and it is the main reason we are writing this up. The 414-row table that all our models were trained on contains 46 distinct yield values, each repeated nine times. The repetition is an artefact of how a temperature file was joined, not of anything that happens in a field. A random split puts copies of the same crop-year on both sides of the train/test boundary, and the model is rewarded for remembering them. Once we evaluated by whole years, the picture was very different, and a one-line linear trend turned out to be the strongest model we had.

We think this is worth reporting for two reasons. First, the same mistake is easy to make with other public agricultural tables, and the cost of making it is a headline number that means little. Second, the prototype itself, including the leaf classifier and the explanation interface, is a reasonable starting point for people working on similar problems in Pakistan, provided its limits are stated plainly. Our contributions are:
\begin{itemize}
  \item A data audit of a public Kaggle-derived crop table, showing that 414 rows correspond to 46 independent observations and that the national pesticide series is an exact straight line in time from 2001 to 2013, with a reproducible deduplication step.
  \item A comparison of Random Forest, XGBoost, SVR, a stacking ensemble and simple baselines under three validation protocols (random, leave-one-year-out and forward chaining), with bootstrap confidence intervals over years, and a test of whether weather and input variables explain what a linear trend misses.
  \item An independent check on a district-level wheat panel for Punjab (36 districts, 13 seasons), which shows how much of the variation is spatial and how much a season-wide shock could add.
  \item A replication on a longer national table that we rebuilt from FAOSTAT (1990--2024, 70 observations), in which the pesticide series, the one feature that seemed to help, turns out to be largely imputed.
  \item A MobileNetV2 leaf-health classifier (99.87\% accuracy on 772 held-out PlantVillage images, 96.4\% recall on 336 unseen diseased leaves after removing near-duplicates of its training images) and a deployed Gradio prototype with SHAP explanations.
  \item A frank account of an integration that did not work: the image-derived feature has to be constructed when no paired data exist, and it adds nothing. We list what data would be needed to test the idea properly.
\end{itemize}

\section{Related Work}
\paragraph{Yield prediction with machine learning.}
A systematic review of the area \citep{vanklompenburg2020} found that temperature, rainfall and soil type are the most common inputs and that deep learning models are widely used alongside classical learners. Statistical yield models are usually built on weather variables, and temperature effects in particular are strongly nonlinear \citep{schlenker2009,lobell2010}. We use three widely used learners: Random Forest \citep{breiman2001}, gradient-boosted trees as implemented in XGBoost \citep{chen2016}, and support vector regression \citep{drucker1997}. We combine them by stacked generalisation \citep{wolpert1992} with a Ridge meta-learner \citep{hoerl1970}.

\paragraph{Plant disease from images.}
Image-based disease recognition has been studied extensively on the PlantVillage collection \citep{hughes2015,mohanty2016}, where deep networks reach very high accuracy on leaves photographed against plain backgrounds. Such accuracy is known to be fragile: models can exploit background cues instead of symptoms \citep{geirhos2020}, and the variety of the training images matters as much as their number \citep{barbedo2018}. We use MobileNetV2 \citep{sandler2018}, chosen because it is small enough to run on a free hosting tier.

\paragraph{Validation with dependent data.}
It has long been known that standard cross-validation assumes exchangeable samples and can give misleading estimates when data have temporal, spatial or hierarchical dependence \citep{roberts2017}, and that evaluating forecasting models needs its own care \citep{bergmeir2012,tashman2000}. A comprehensive survey of cross-validation procedures is given by \citet{arlot2010}. In the forecasting literature, simple benchmark methods are repeatedly found hard to beat by more complex ones \citep{makridakis2018}, which motivates the baselines we use. Leakage of this kind is a recognised source of over-optimistic results across many fields \citep{kaufman2012,kapoor2023}. What we add here is a concrete and small example, from a public table, of how a join can create the leak without anyone intending it.

\section{Data}
\label{sec:data}
We used four public datasets. Table~\ref{tab:data} summarises them; Figure~\ref{fig:pipeline} shows how they are processed.

\begin{table}[H]
\centering
\caption{Datasets used. The first two feed the national yield models; the FAOSTAT table and the district panel are used for the replications in Sections~\ref{sec:faoresults} and~\ref{sec:district}; the corn dataset is used only to test the classifier on unseen images; the wheat statistics were used for exploratory plots.}
\label{tab:data}
\small
\begin{tabular}{p{3.4cm}p{2.0cm}p{6.8cm}}
\toprule
Dataset & Size & Content used \\
\midrule
Crop Yield Prediction \citep{patel2022} & 28{,}242 rows (1{,}449 for Pakistan) & FAO yields (hg/ha), pesticide use (tonnes), national mean temperature, 1990--2013 \\
District-level crop data, Punjab (Kaggle) & 2{,}200 rows, 13 crops & Soil N, P, K, soil temperature, humidity, pH, rainfall \\
Wheat statistics, Punjab (Kaggle) & 12 years & Area, production, yield (trend plots only) \\
District wheat yields, Punjab \citep{pbsdistrict} & 36 districts $\times$ 13 seasons & Wheat yield (kg/ha), 2007--08 to 2019--20 (independent check) \\
FAOSTAT, re-downloaded \citep{faostat} & 35 years $\times$ 2 crops & Official yields, pesticide use, temperature change, 1990--2024 (replication) \\
PlantVillage \citep{plantvillagekaggle,hughes2015} & 3{,}852 maize images & Healthy, common rust, northern leaf blight, gray leaf spot \\
Corn or Maize Leaf Disease \citep{cornkaggle} & 4{,}188 images & Candidate images for the unseen-image test \\
\bottomrule
\end{tabular}
\end{table}

\subsection{Tabular data and the merged table}
We kept the Pakistan rows of the yield table (whose yield and pesticide columns, according to the dataset description, come from FAO statistics \citep{faostat}), converted yield from hg/ha to Ton/Ha (division by $10^4$), and averaged the district-level soil and climate values by crop. The two tables share only wheat and maize, so an inner join on crop type left 414 rows covering 1990--2013. There are 23 distinct years, because 2003 is absent. We also derived \code{NPK\_total} $= N + P + K$. The merged table has no missing values.

Two properties of this table matter later. First, the soil variables (N, P, K, soil temperature, humidity, pH) and rainfall (a constant 494 mm) do not vary within a crop, because they come from crop-level averages. They therefore act as a crop indicator and nothing more. Second, pesticide use and temperature are national values that change from year to year but are shared by the two crops.

\paragraph{The pesticide series.} Pesticide use is a single national figure per year, shared by both crops. Looking at it over time (Figure~\ref{fig:audit}d), we noticed that from 2001 to 2013 the values fall by the same 989.2 tonnes every year, from 11{,}871 tonnes to under one tonne, and then stay constant at 0.92 tonnes in the later years of the source file. The same straight line appears in the original pesticide file before any merging. Real application figures do not usually behave like this; the values look interpolated or imputed, although we could not trace this in the source documentation. The consequence is that for 13 of the 23 years the pesticide column is simply a linear function of the year, and, as Sections~\ref{sec:honest} and~\ref{sec:detrend} show, models use it as one. Section~\ref{sec:faodata} shows that the current FAOSTAT release gives a different series, which is mostly imputed.

\subsection{The nine-fold replication}
\label{sec:audit}
\begin{figure}[H]
\centering
\includegraphics[width=0.92\textwidth]{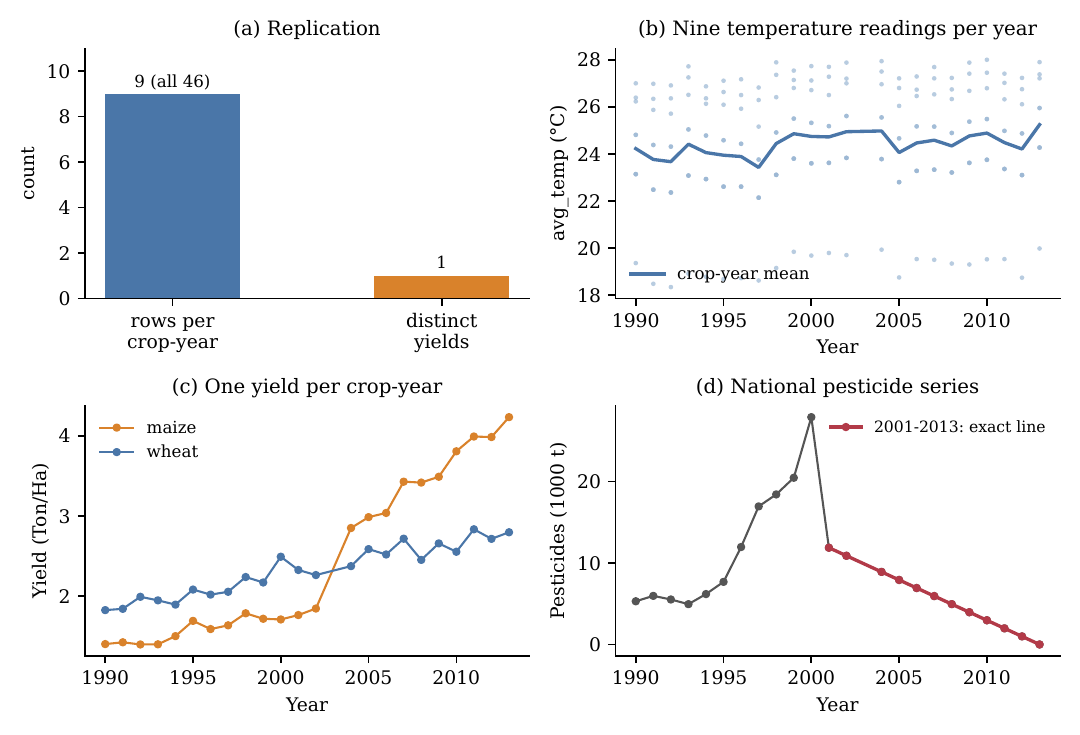}
\caption{Data audit. (a) Every crop-year appears nine times with a single yield. (b) The nine temperature readings per year (light dots) vary much more within a year than the yearly mean (line) varies across years; shown for wheat. (c) Yield by crop and year. Both series rise steadily and correlate with year at $r = 0.95$ (maize) and $r = 0.96$ (wheat). (d) The national pesticide series: from 2001 it is an exact straight line, falling by 989.2 tonnes per year to under one tonne.}
\label{fig:audit}
\end{figure}

Each Pakistan crop-year appears exactly nine times in the merged table, always with the same yield and the same pesticide figure, but with different temperatures (Figure~\ref{fig:audit}a,b). The cause is the temperature source file, which holds nine readings for each Pakistan year. Their provenance is not documented in the files we used (they could be station-level or regional values), and their spread within a year (mean standard deviation 2.5\,$^\circ$C) is five times larger than the spread of the yearly means (0.5\,$^\circ$C). A join on country and year therefore multiplied every crop-year row by nine. The 414 rows are $2 \text{ crops} \times 23 \text{ years} \times 9$; the independent information is in 46 rows, each with one yield.

We kept the 414-row table for the deployed models, so that the prototype matches our original submission. For every honest evaluation in this paper we collapse the replicates into one row per crop-year, taking the mean temperature. We call this the \emph{crop-year table}. It is released with the code.

\subsection{District-level wheat panel}
\label{sec:districtdata}
To test our conclusions on data with real spatial variation, we added a second, independent source: wheat yields per hectare by district for Punjab, from 2007--08 to 2019--20, published by the Pakistan Bureau of Statistics and redistributed through the Open Data Pakistan portal \citep{pbsdistrict}. We removed the province and division summary rows and the row for Islamabad, which lies outside Punjab. Chiniot has no values for the first two seasons (the district was created later) and was kept for the remaining eleven. Two cells were removed as implausible: Kasur in 2009--10 (8.9 Ton/Ha) and Lahore in the same season (0.78 Ton/Ha), whose values differ from the reported figure for their division (2.61 Ton/Ha) by a factor of three or more; both look like transcription errors, and no other cell deviates from its division by more than 60\%. The resulting panel has 464 observations from 36 districts and 13 seasons. Seasons are labelled by harvest year. As a consistency check of the earlier national table, the province-level row of this file correlates at $r = 0.99$ with the national FAO wheat yields in our table for the six overlapping harvest years (2008--2013). The portal lists no licence for the file, so we ship the cleaning code and not the raw data (see Data and Code Availability). This panel contains yields only; it has no weather variables.

\subsection{FAOSTAT table, 1990--2024}
\label{sec:faodata}
To check our conclusions on data that we could trace to its source, we re-downloaded Pakistan's wheat and maize yields, national pesticide use and national temperature change from FAOSTAT \citep{faostat} and built a table for 1990--2024: 35 years and two crops, 70 crop-year observations, with no gap in 2003. The 46 crop-years that overlap with the Kaggle-derived table have identical yields (the largest difference is below $10^{-15}$ Ton/Ha), so the yield column of that table is a faithful copy of FAO figures. All yields are official figures except maize in 2024, which FAOSTAT flags as a figure from an international organisation. Temperature enters as the annual national temperature change in $^\circ$C (meteorological year); the yearly mean temperature in the Kaggle table correlates with it at $r=0.92$ over the 23 overlapping years, so the two describe the same signal at national scale.

\begin{figure}[H]
\centering
\includegraphics[width=\textwidth]{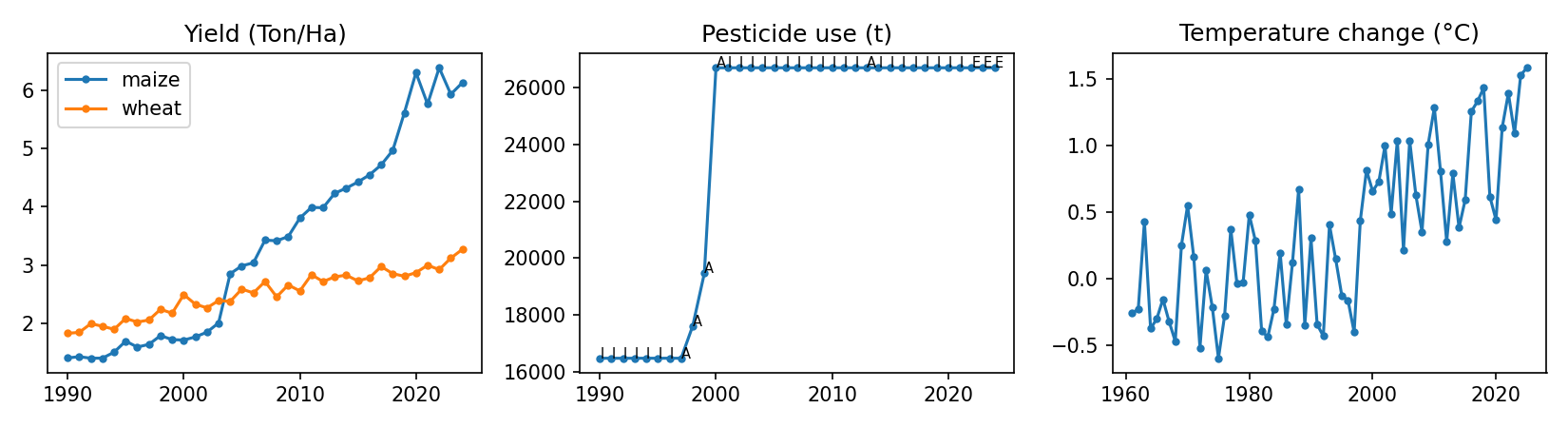}
\caption{The FAOSTAT table. Left: yields of maize and wheat, 1990--2024. Middle: national pesticide use with FAOSTAT's flag for each year (A official, E estimated, I imputed); the series is flat except for 1997--2000. Right: annual national temperature change (the panel shows the full series from 1961).}
\label{fig:fao}
\end{figure}

Pesticide use is a different matter (Figure~\ref{fig:fao}). The current FAOSTAT series takes only four distinct values: 16{,}479 t in 1990--1997, 17{,}619 t in 1998, 19{,}470 t in 1999 and 26{,}702 t from 2000 to 2024. FAOSTAT's own flags mark 27 of the 35 values as imputed, three (2022--2024) as estimated and only five (1997--2000 and 2013) as official. This series does not resemble the one in the Kaggle table (correlation $-0.25$ over the overlapping years): there the values start at 5{,}299 t in 1990, peak in 2000 and then fall in a straight line to 0.92 t in 2013. We did not establish whether the two come from different FAO releases. Either way, neither looks like measured annual pesticide use.

\subsection{Leaf images}
From PlantVillage we used the four maize folders. The three disease classes (common rust, northern leaf blight, gray leaf spot) were merged into one \emph{diseased} label, giving a binary problem with 2{,}690 diseased and 1{,}162 healthy images. Wheat leaf classes are not part of PlantVillage, so the image module covers maize only.

\section{Methodology}
\subsection{Preprocessing pipeline}
\begin{figure}[H]
\centering
\includegraphics[width=\textwidth]{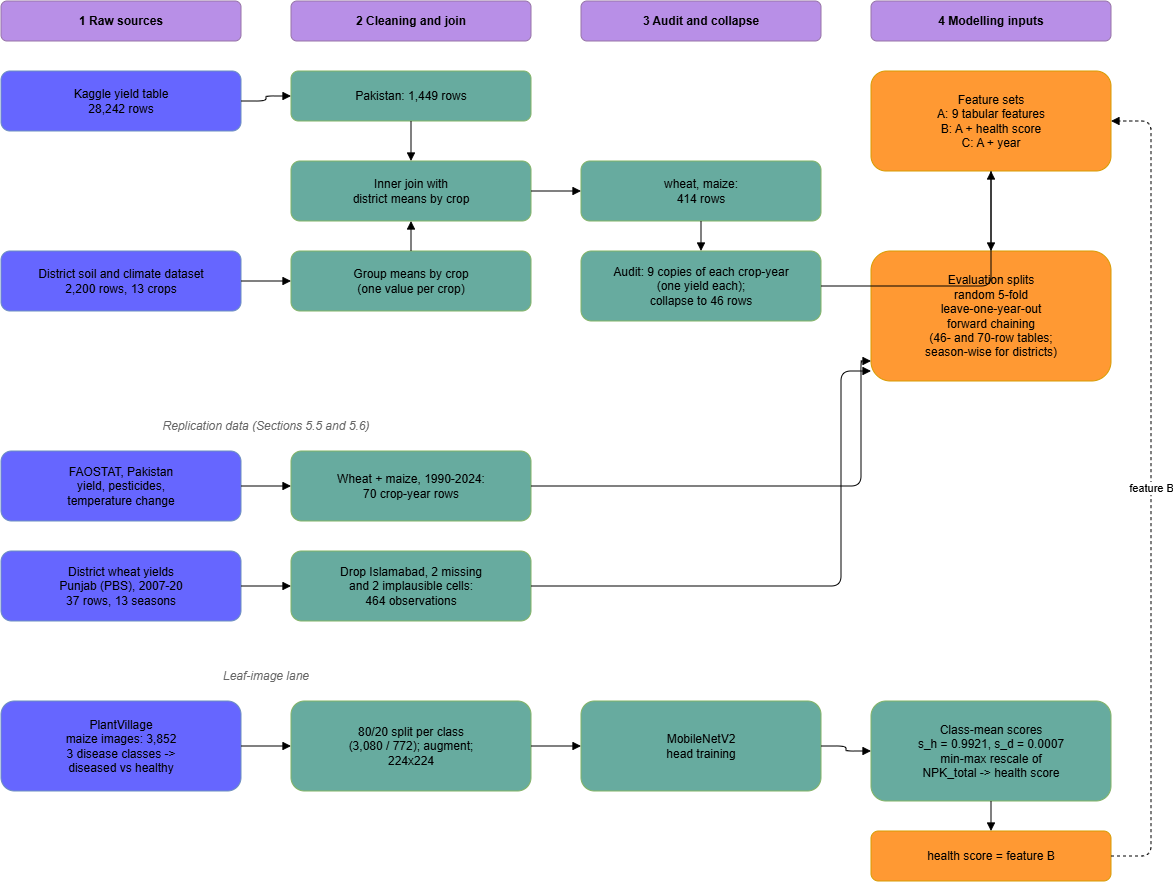}
\caption{Data pipeline. Blue: raw sources; green: cleaning, join and audit steps; orange: modelling inputs and evaluation; purple: stage headers. The 414-row table produced by the join trains the deployed models; the 46-row table produced by the audit is used for the honest-protocol evaluation of the Kaggle-derived data. The FAOSTAT table and the district wheat panel (middle rows) are used for replication, and the lower lane trains the leaf classifier whose score feeds feature set B. The 12-year Punjab wheat statistics were used for exploratory plots only.}
\label{fig:pipeline}
\end{figure}

Figure~\ref{fig:pipeline} summarises the full pipeline. The tabular lane (blue) goes from the raw yield file through the Pakistan filter, the district-level aggregation, the join and the audit to the three feature sets described below. The image lane (orange) trains the leaf classifier and produces the score that feeds feature set B. The two teal lanes show the FAOSTAT table and the district wheat panel that we use for replication. The only scaling step is a standardisation inside the SVR and Ridge pipelines; tree models receive raw values. All random seeds are fixed at 42.

\subsection{Feature sets}
We use three feature sets throughout.
\begin{description}[leftmargin=1.2cm,labelwidth=1cm]
  \item[A] Nine tabular features: pesticide use, mean temperature, N, P, K, soil temperature, humidity, pH and \code{NPK\_total}. This is the set used by the deployed models.
  \item[B] Set A plus \code{health\_score}, the output of the leaf classifier (Section~\ref{sec:integration}).
  \item[C] Set A plus calendar year. Year is known in advance, so it is a legitimate input for forecasting; we add it as a separate set to see what the models do when given the one variable that tracks the trend.
\end{description}
An earlier 13-feature version also contained year, rainfall, a temperature--rainfall product and a crop code. We dropped rainfall (constant), the product and the crop code for being redundant, and year because we wanted the models to rely on weather, input and soil information. As Section~\ref{sec:results} shows, that choice turned out to matter a great deal.

\subsection{Regression models}
Let $\mathbf{x}_i$ be the features of crop-year $i$ and $y_i$ its yield in Ton/Ha. We fit five regressors.

\paragraph{Baselines.} The \emph{crop mean} predicts the training mean yield of the crop. The \emph{crop linear trend} fits $y = a_c + b_c\,t$ separately for each crop $c$ by ordinary least squares on year $t$. Both use no weather or soil information.

\paragraph{Ridge regression.} A linear model with penalty $\lambda\lVert\boldsymbol\beta\rVert^2$, $\lambda=1$, on standardised inputs.

All learners are implemented in scikit-learn \citep{pedregosa2011} or XGBoost \citep{chen2016}.

\paragraph{Random Forest} with 200 trees and maximum depth 8, \textbf{XGBoost} with 200 trees, learning rate 0.05 and maximum depth 5, and \textbf{SVR} with an RBF kernel ($C=100$, $\gamma=0.1$, $\epsilon=0.1$) on standardised inputs. These settings were chosen once, early in the project, and were not tuned; we use the same settings under every protocol. (The earlier 13-feature experiments used depth 10 and 6 for the two tree models.)

\paragraph{Stacking ensemble.} The three learners above are first-level models. A Ridge regression ($\lambda=1$) is the second-level model and is fitted on out-of-fold predictions from five-fold cross-validation:
\begin{equation}
\hat y(\mathbf{x}) \;=\; \beta_0 + \sum_{k\in\{\text{RF},\text{XGB},\text{SVR}\}} \beta_k\, \hat f_k(\mathbf{x}), \qquad
\boldsymbol\beta = \arg\min_{\boldsymbol\beta}\ \sum_i \Big(y_i - \beta_0 - \sum_k \beta_k \tilde f_k(\mathbf{x}_i)\Big)^2 + \lambda\lVert\boldsymbol\beta\rVert^2,
\end{equation}
where $\tilde f_k$ denotes the out-of-fold prediction. In the deployed model trained on 414 rows, the learned weights were $\beta_{\text{XGB}}=0.967$, $\beta_{\text{RF}}=0.070$ and $\beta_{\text{SVR}}=-0.027$, with an intercept of $-0.021$. The ensemble is therefore close to XGBoost alone.

\subsection{Leaf-health classifier}
\label{sec:cv}
\begin{figure}[H]
\centering
\includegraphics[width=\textwidth]{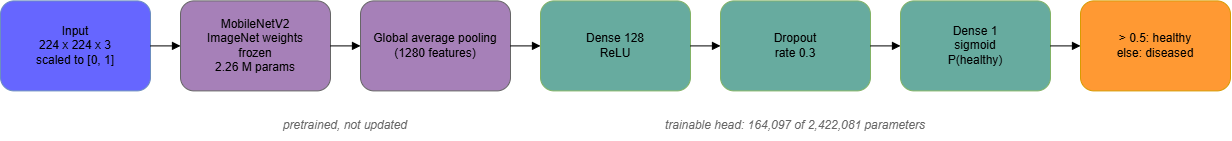}
\caption{Architecture of the leaf-health classifier. Only the dense head is trained; the pretrained backbone is kept frozen.}
\label{fig:cvarch}
\end{figure}

The classifier (Figure~\ref{fig:cvarch}) is a MobileNetV2 \citep{sandler2018} pretrained on ImageNet, with its convolutional backbone frozen. On top we train a small head: global average pooling, a dense layer of 128 ReLU units, dropout with rate 0.3, and one sigmoid unit that outputs $P(\text{healthy})$. Of 2{,}422{,}081 parameters, 164{,}097 are trainable. Images are resized to $224\times224$ and scaled to $[0,1]$. Each class folder was split 80/20 with a fixed seed, giving 3{,}080 training and 772 validation images (539 diseased, 233 healthy). Training images were augmented with rotations up to $20^\circ$, zoom up to 20\%, horizontal flips and brightness changes between 0.8 and 1.2. We minimised binary cross-entropy with Adam, monitored validation accuracy, stopped after three epochs without improvement, and kept the best weights. Training ended after five epochs. A cut-off of 0.5 turns the probability into a label.

\subsection{Integrating the leaf score into the yield models}
\label{sec:integration}
At inference time the integration is straightforward: the uploaded photograph is scored by the classifier and the probability becomes \code{health\_score}; if no photograph is uploaded, a neutral value of 0.5 is used. At training time there is a difficulty. Our yield records are national, annual figures and carry no leaf images. To give the yield models something to learn from, we built a score for each training row. Let $s_h = 0.9921$ and $s_d = 0.0007$ be the classifier's mean predicted probabilities on validation leaves of the healthy and diseased classes. For row $i$ we set
\begin{equation}
\code{health\_score}_i \;=\; s_d + \frac{\code{NPK\_total}_i - \min_j \code{NPK\_total}_j}{\max_j \code{NPK\_total}_j - \min_j \code{NPK\_total}_j}\,(s_h - s_d).
\end{equation}
This is a rescaled copy of an existing feature. Because \code{NPK\_total} is constant within a crop (159.63 for wheat, 160.25 for maize), the score takes only two values, 0.0007 for wheat rows and 0.9921 for maize rows. Wheat is scored near zero only because its nutrient total happens to be slightly lower; no wheat image was involved, and the value does not mean that wheat is unhealthy. We disclose this construction in full because it limits what the integration can show. We test its effect in Section~\ref{sec:health}.

\subsection{Evaluation of the classifier on unseen images}
\label{sec:unseenmethods}
The PlantVillage validation set comes from the same collection as the training set, so it says little about other image sources. To test on other images we used the Kaggle \emph{Corn or Maize Leaf Disease Dataset} \citep{cornkaggle}, which according to its description was curated from PlantVillage and PlantDoc \citep{singh2020} and holds 4{,}188 images in four folders (healthy, blight, common rust, gray leaf spot). Most of those images are PlantVillage images that the classifier was trained on, so we removed every image that is a near-duplicate of a PlantVillage maize image. For each image we computed a 64-bit perceptual hash (the $8\times8$ DCT hash of the image resized to $128\times128$) and treated two images as the same picture when their hashes differed in at most six bits. All 3{,}852 PlantVillage maize images were hashed. Of the 4{,}188 candidates, 3{,}852 matched a PlantVillage image exactly (distance 0), and the other 336 were at distance 12 or more, so the split is clean. The matched images have the same class counts as the PlantVillage maize folders (1{,}162 healthy, 1{,}192 common rust, 985 blight, 513 gray leaf spot); the 336 others (161 blight, 114 common rust, 61 gray leaf spot) are the images that PlantVillage does not account for. We take them as unseen images. Their source is presumably PlantDoc, but we did not check provenance image by image. All of them are diseased, because the healthy folder consists entirely of PlantVillage images. We therefore report the recall of the diseased class with a Wilson interval, and could not measure the false-alarm rate on healthy leaves. As a contamination control we also evaluated a random sample of 800 of the matched images.

\subsection{Evaluation protocols}
\label{sec:protocols}
\begin{figure}[H]
\centering
\includegraphics[width=0.92\textwidth]{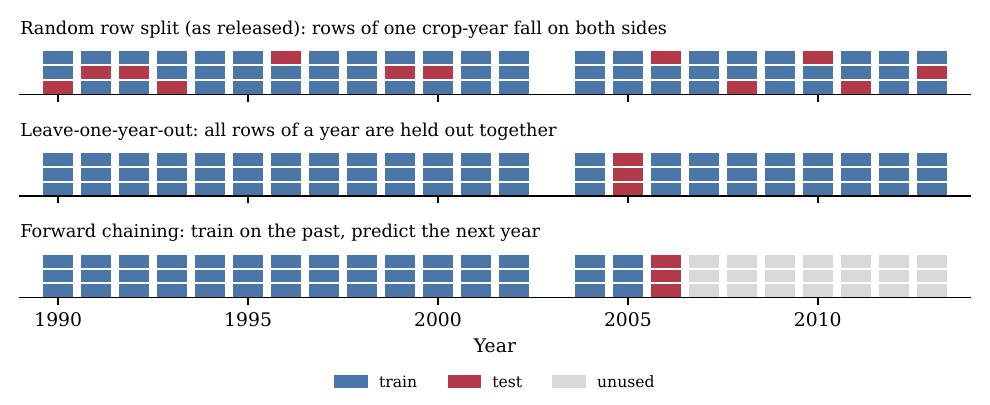}
\caption{The three validation protocols. Each small block is one row (three rows per year are drawn for illustration; 2003 is absent from the data). In a random split, rows of the same crop-year land on both sides; in the other two, a year is either fully held out or lies in the future of the training data.}
\label{fig:protocols}
\end{figure}

We compare three ways of splitting the data (Figure~\ref{fig:protocols}).
\begin{itemize}
  \item \textbf{Random 5-fold.} Rows are shuffled and split into five folds, repeated five times. This is the conventional protocol and matches how our original results were produced.
  \item \textbf{Leave-one-year-out (LOYO)}, a blocked form of cross-validation \citep{roberts2017,arlot2010}. All rows of one year, both crops, are held out together; 23 folds.
  \item \textbf{Forward chaining.} For each test year from 1998 to 2013 the models are trained on all earlier years only (15 test years). This mimics forecasting and is also known as rolling-origin evaluation \citep{tashman2000,hyndman2021}.
\end{itemize}
Both crops of a year share the same pesticide and temperature values, so holding out only one of them would leave its twin in the training set. For this reason we group by year and not by crop-year.

We report RMSE \citep{hyndman2006}, MAE and $R^2$ computed on the pooled held-out predictions. For LOYO and forward chaining we also give a 95\% bootstrap interval for the RMSE, obtained by resampling the held-out years with replacement 2{,}000 times \citep{efron1993}. With 15--23 years these intervals are wide, and we ask the reader to take them seriously. For the leaf classifier we report accuracy with a Wilson score interval \citep{wilson1927}.

\subsection{District-panel evaluation}
\label{sec:districtmethods}
On the district panel (Section~\ref{sec:districtdata}) we compare six predictors of the yield of district $d$ in season $t$. The \emph{district mean} is the average yield of that district in the training seasons. The \emph{common trend} adds a single linear trend in season index, fitted to the training residuals of all districts. Ridge regression, Random Forest and XGBoost use one-hot district indicators and the season index as inputs, with the same settings as before. The \emph{oracle season shock} adds to the district mean the average deviation of all districts in the held-out season, computed from that season's own labels. It cannot be used in practice; its purpose is to measure how much a perfect season-wide predictor, for example one built from weather, could add. We evaluate by random 5-fold (three repeats), leave-one-season-out, and forward chaining with test seasons from 2012--13 onward (training on at least five earlier seasons), with bootstrap intervals over test seasons.

\subsection{Replication on the FAOSTAT table}
\label{sec:faomethods}
On the FAOSTAT table (Section~\ref{sec:faodata}) we repeat the comparison of trend and models with two feature sets: A$'$ (pesticide use, temperature change and a crop indicator) and C$'$ (A$'$ plus year). The soil variables are left out because they are crop-level constants that duplicate the crop indicator. The models, settings, protocols and bootstrap are those of Sections~\ref{sec:protocols} and~\ref{sec:detrend}, with 35 leave-one-year-out folds and forward chaining over the 27 test years 1998--2024. The residual models use the same small settings as before (Section~\ref{sec:hyper}).

\subsection{Explainability and deployment}
\begin{figure}[H]
\centering
\includegraphics[width=\textwidth]{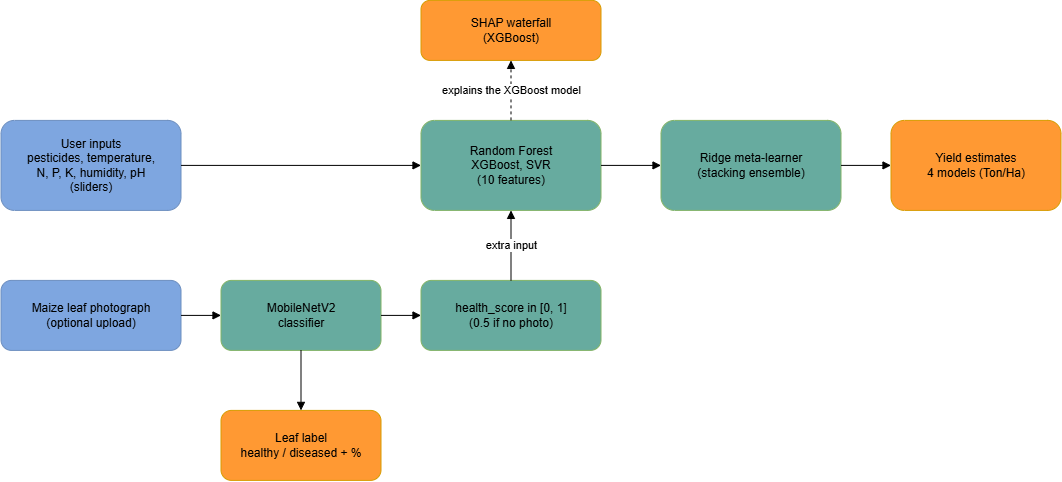}
\caption{Overview of the deployed prototype. A user sets climate and soil sliders and may upload a maize leaf photograph. The classifier output enters the yield models as an extra feature (0.5 if no photograph is uploaded). The SHAP plot explains the XGBoost model; the yield estimates come from all four models.}
\label{fig:overview}
\end{figure}

Predictions of the XGBoost model are explained with SHAP \citep{lundberg2017} (TreeExplainer), displayed as a waterfall plot. The application (Figure~\ref{fig:overview}) is written with Gradio \citep{abid2019}, uses scikit-learn \citep{pedregosa2011}, XGBoost and Keras \citep{chollet2015}, and is hosted on Hugging Face Spaces. Users set sliders for pesticide use (0--30{,}000\,t), mean temperature (15--35\,$^\circ$C), N (0--140), P (0--100), K (0--120), soil temperature, humidity (50--100\%) and pH (4--9), optionally upload a maize leaf image, and receive the four model predictions, a comparison chart, the leaf label and the SHAP plot.

\section{Results}
\label{sec:results}

\subsection{Results as originally obtained}
On a random 80/20 split of the 414-row table (seed 42), the deployed models reached the scores in Table~\ref{tab:original}. The stacking ensemble led with $R^2=0.9906$ and a mean absolute error of 0.046 Ton/Ha, XGBoost followed closely, and SVR was poor. With the earlier 13-feature set, which included year, Random Forest, XGBoost and the ensemble were essentially perfect ($R^2 \geq 0.9999$). Taken at face value, this is an extremely accurate system.

\begin{table}[H]
\centering
\caption{Original evaluation: random 80/20 split of the 414-row table, nine features plus \code{health\_score} (feature set B). \textbf{These figures are inflated by the replication discussed in Section~\ref{sec:audit}.}}
\label{tab:original}
\small
\begin{tabular}{lcccc}
\toprule
Model & MAE & RMSE & Test $R^2$ & 5-fold CV $R^2$ \\
\midrule
Random Forest & 0.0715 & 0.1309 & 0.9735 & 0.9582 \\
XGBoost & 0.0509 & 0.0827 & 0.9894 & 0.9922 \\
SVR & 0.4639 & 0.7086 & 0.2240 & 0.3767 \\
Stacking ensemble & 0.0462 & 0.0780 & 0.9906 & 0.9923 \\
\bottomrule
\end{tabular}
\end{table}

\subsection{The same rows under a different split}
\label{sec:leak}
\begin{figure}[H]
\centering
\includegraphics[width=\textwidth]{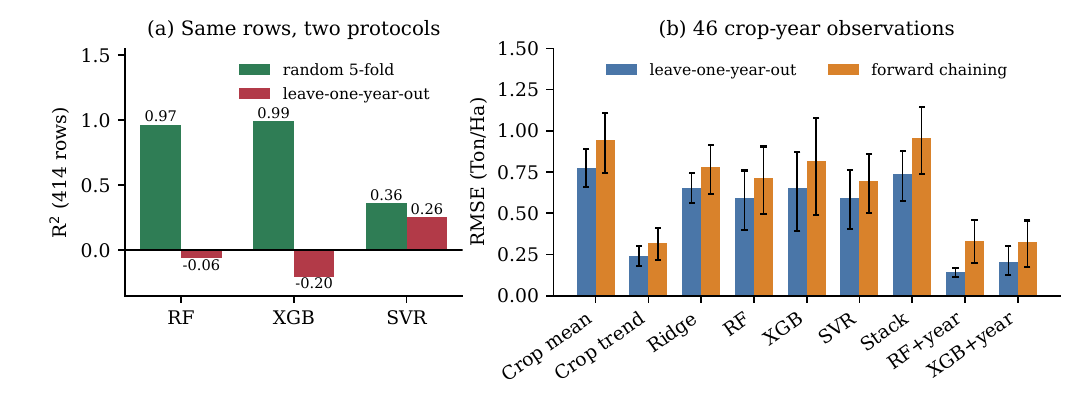}
\caption{(a) $R^2$ of three models on the 414-row table under random 5-fold and leave-one-year-out validation. (b) Pooled RMSE on the 46-row crop-year table under leave-one-year-out and forward chaining, with 95\% bootstrap intervals over years. ``+year'' means feature set C.}
\label{fig:results}
\end{figure}

We re-evaluated the same 414 rows and the same feature set A (no leaf score), changing only how rows are assigned to folds. Table~\ref{tab:leak} and Figure~\ref{fig:results}a show that the result depends almost entirely on the protocol. Under random 5-fold, XGBoost reaches $R^2=0.994$. When the nine copies of a year are kept together (LOYO), the same model falls to $R^2=-0.204$, which is worse than predicting the overall mean. Random Forest drops from 0.965 to $-0.063$. SVR is the exception: it was poor to begin with and changes less, because its smooth kernel cannot memorise individual crop-years as easily.

\begin{table}[H]
\centering
\caption{Same 414 rows, same features (set A), two protocols. Random 5-fold is a single shuffled run.}
\label{tab:leak}
\small
\begin{tabular}{lrrrr}\toprule
Model & \multicolumn{2}{c}{Random 5-fold} & \multicolumn{2}{c}{Leave-one-year-out} \\\cmidrule(lr){2-3}\cmidrule(lr){4-5}
 & RMSE & R$^2$ & RMSE & R$^2$ \\\midrule
Random Forest & 0.139 & 0.965 & 0.768 & -0.063 \\
XGBoost & 0.060 & 0.994 & 0.818 & -0.204 \\
SVR & 0.597 & 0.359 & 0.643 & 0.255 \\
\bottomrule\end{tabular}
\end{table}

\subsection{Honest evaluation on the crop-year table}
\label{sec:honest}
Table~\ref{tab:protocols} gives results on the 46-row crop-year table for the three protocols; Figure~\ref{fig:results}b plots the two year-aware ones. Five observations stand out.

\begin{table}[H]
\centering
\caption{Performance on the 46-row crop-year table. RMSE in Ton/Ha; for the two year-aware protocols the 95\% bootstrap interval over held-out years is shown in brackets. Random 5-fold is the mean of five repeats. Feature set A has no year; set C adds year.}
\label{tab:protocols}
\footnotesize
\setlength{\tabcolsep}{4pt}
\begin{tabular}{lrrrrrr}\toprule
 & \multicolumn{2}{c}{Random 5-fold} & \multicolumn{2}{c}{Leave-one-year-out} & \multicolumn{2}{c}{Forward chaining} \\
\cmidrule(lr){2-3}\cmidrule(lr){4-5}\cmidrule(lr){6-7}
Model & RMSE & R$^2$ & RMSE [95\% CI] & R$^2$ & RMSE [95\% CI] & R$^2$ \\\midrule
\multicolumn{7}{l}{\emph{Baselines}} \\
Crop mean & 0.779 & -0.09 & 0.777 \,[0.66, 0.89] & -0.09 & 0.943 \,[0.75, 1.11] & -0.87 \\
Crop linear trend & 0.240 & 0.90 & 0.243 \,[0.18, 0.30] & 0.89 & 0.317 \,[0.22, 0.41] & 0.79 \\
\midrule\multicolumn{7}{l}{\emph{Feature set A (nine features, no year)}} \\
Ridge & 0.651 & 0.24 & 0.654 \,[0.56, 0.74] & 0.23 & 0.779 \,[0.62, 0.91] & -0.28 \\
Random Forest & 0.516 & 0.52 & 0.593 \,[0.40, 0.76] & 0.37 & 0.713 \,[0.50, 0.91] & -0.07 \\
XGBoost & 0.503 & 0.54 & 0.651 \,[0.39, 0.87] & 0.24 & 0.817 \,[0.49, 1.08] & -0.40 \\
SVR & 0.586 & 0.38 & 0.592 \,[0.41, 0.76] & 0.37 & 0.695 \,[0.50, 0.86] & -0.01 \\
Stacking ensemble & 0.748 & -0.01 & 0.737 \,[0.57, 0.88] & 0.02 & 0.956 \,[0.74, 1.14] & -0.92 \\
\midrule\multicolumn{7}{l}{\emph{Feature set C (A + year)}} \\
Ridge + year & 0.436 & 0.66 & 0.407 \,[0.35, 0.46] & 0.70 & 0.507 \,[0.40, 0.59] & 0.46 \\
Random Forest + year & 0.168 & 0.95 & 0.141 \,[0.11, 0.17] & 0.96 & 0.333 \,[0.20, 0.46] & 0.77 \\
XGBoost + year & 0.211 & 0.92 & 0.207 \,[0.12, 0.30] & 0.92 & 0.323 \,[0.17, 0.46] & 0.78 \\
SVR + year & 0.271 & 0.86 & 0.303 \,[0.18, 0.43] & 0.83 & 0.354 \,[0.24, 0.45] & 0.74 \\
Stacking ensemble + year & 0.272 & 0.86 & 0.275 \,[0.19, 0.36] & 0.86 & 0.388 \,[0.28, 0.50] & 0.68 \\
\bottomrule\end{tabular}
\end{table}

\paragraph{A one-line trend beats every model without year.}
The per-crop linear trend gives an RMSE of 0.243 Ton/Ha under LOYO (interval 0.18--0.30) and 0.317 under forward chaining. Every learned model that uses feature set A lies between 0.59 and 0.74 under LOYO and between 0.70 and 0.96 under forward chaining, and the intervals do not overlap with the trend baseline's. The crop mean, which uses nothing beyond the crop, scores 0.78 and 0.94. The models built on weather, pesticide and soil features do better than that, but only modestly under LOYO, and under forward chaining every one of them has a negative $R^2$: they forecast worse than a constant equal to the mean of the test years, and the ensemble (0.956) is even slightly worse than the crop mean (0.943). The signal in our table is overwhelmingly a time trend: yield correlates with year at $r\approx0.95$ in both crops (Figure~\ref{fig:audit}c), which reflects improving varieties, area and practices that none of our features describe.

\begin{figure}[H]
\centering
\includegraphics[width=\textwidth]{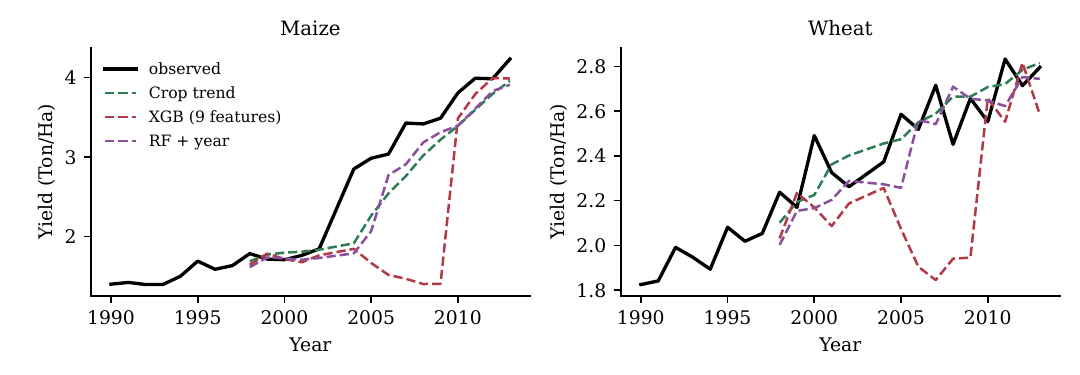}
\caption{Forward-chaining predictions for each test year (trained on all earlier years). The linear trend and the year-aware random forest follow the rising yields; the nine-feature XGBoost model does not.}
\label{fig:forward}
\end{figure}

\paragraph{Giving the models the year changes the picture.}
With year added (set C), Random Forest reaches an LOYO RMSE of 0.141 ($R^2=0.964$, interval 0.11--0.17), clearly better than the linear trend, and XGBoost reaches 0.207. This is interpolation: a held-out year sits between years the model has seen, and a tree can learn a step-wise trend. Forward chaining is harder because it requires extrapolation, which tree models cannot do, since they predict piecewise-constant values \citep{hastie2009}. There the advantage disappears: RF and XGBoost with year give 0.333 and 0.323, against 0.317 for the plain trend, a tie within the intervals (Figure~\ref{fig:forward}). Models without year fail more visibly: the nine-feature XGBoost forecast falls while observed yields rise.

\paragraph{Replication, not noise, produced the headline numbers.}
Under random 5-fold on the crop-year table, XGBoost with set A reaches $R^2 = 0.54$ rather than 0.994 (Table~\ref{tab:protocols}, compared with Table~\ref{tab:leak}). The protocol is the same; the difference is that the nine copies are gone. This isolates the replication as the main source of inflation, ahead of the choice of model.

\paragraph{The ensemble is not better than its parts here.}
The stacking ensemble is the best model in the original evaluation but is worse than RF, XGBoost and SVR individually under the honest protocols (LOYO RMSE 0.737, set A). We have not investigated this fully. Our guess is that with 45 or fewer training rows the inner five-fold cross-validation that produces the second-level training data is itself unstable, but we did not test this.

\paragraph{A caution on interpreting small intervals.}
The bootstrap intervals are wide, and the number of years is small. We would not rank RF, XGBoost and SVR against one another on these data, and we do not claim that the year-aware forest is better than a trend model for forecasting, only that it is better at interpolation.

\subsection{Modelling deviations from the trend}
\label{sec:detrend}
Since a straight line per crop is so hard to beat, a natural question is whether weather, soil and input variables explain what the line misses. We tested this with a two-stage model. In every fold, the per-crop trend is fitted on the training years only; a regressor is then trained to predict the training residuals from the non-time features (crop indicator, mean temperature, pesticide use); and the prediction for a held-out year is the trend plus the predicted residual. The regressors were kept small (Section~\ref{sec:hyper}) and were chosen once, without tuning. All four residual models match or improve on the trend alone (Table~\ref{tab:detrend}). The forest does best: its leave-one-year-out RMSE is 0.159 against 0.243 for the trend, and its forward-chaining RMSE is 0.260 against 0.317.

\begin{table}[H]
\centering
\caption{Trend plus residual model on the 46-row crop-year table. Residual features: crop indicator, mean temperature and pesticide use.}
\label{tab:detrend}
\small
\begin{tabular}{lcccccc}\toprule
 & \multicolumn{3}{c}{Leave-one-year-out} & \multicolumn{3}{c}{Forward chaining} \\\cmidrule(lr){2-4}\cmidrule(lr){5-7}
Model & RMSE & 95\% CI & R$^2$ & RMSE & 95\% CI & R$^2$ \\\midrule
Crop linear trend (no residual model) & 0.243 & [0.18, 0.30] & 0.89 & 0.317 & [0.22, 0.41] & 0.79 \\
Trend + Ridge residual & 0.224 & [0.16, 0.28] & 0.91 & 0.306 & [0.21, 0.41] & 0.80 \\
Trend + Random Forest residual & 0.159 & [0.12, 0.19] & 0.95 & 0.260 & [0.14, 0.36] & 0.86 \\
Trend + XGBoost residual & 0.175 & [0.12, 0.23] & 0.94 & 0.275 & [0.18, 0.37] & 0.84 \\
Trend + SVR residual & 0.211 & [0.16, 0.26] & 0.92 & 0.296 & [0.17, 0.41] & 0.82 \\
\bottomrule\end{tabular}
\end{table}

To see which input produces the improvement, we refitted the forest on subsets of the features and bootstrapped the paired RMSE difference to the trend over years (Table~\ref{tab:detrend_ab}). A crop indicator alone reproduces the trend exactly, as it should. Mean temperature adds nothing (gain $+0.001$ under LOYO, interval $-0.034$ to $0.027$). Pesticide use produces the entire gain, $0.085$ Ton/Ha under LOYO (interval 0.048--0.117) and $0.069$ under forward chaining (0.038--0.101). Using calendar year instead gives a gain of a similar kind (0.070 and 0.045).

\begin{table}[H]
\centering
\caption{Which residual features matter (Random Forest residual model). Gain is the reduction in RMSE relative to the trend alone, with a paired bootstrap 95\% interval over held-out years.}
\label{tab:detrend_ab}
\footnotesize
\begin{tabular}{lcccc}\toprule
Residual features (RF) & \multicolumn{2}{c}{Leave-one-year-out} & \multicolumn{2}{c}{Forward chaining} \\\cmidrule(lr){2-3}\cmidrule(lr){4-5}
 & RMSE & gain over trend [95\% CI] & RMSE & gain over trend [95\% CI] \\\midrule
Crop indicator only & 0.243 & +0.000 [-0.001, +0.001] & 0.316 & +0.001 [-0.000, +0.002] \\
+ year (for comparison) & 0.173 & +0.070 [+0.044, +0.093] & 0.274 & +0.045 [+0.019, +0.074] \\
+ mean temperature & 0.243 & +0.001 [-0.034, +0.027] & 0.329 & -0.012 [-0.047, +0.018] \\
+ pesticide use & 0.158 & +0.085 [+0.048, +0.117] & 0.251 & +0.069 [+0.038, +0.101] \\
+ temperature + pesticides & 0.159 & +0.084 [+0.050, +0.115] & 0.260 & +0.060 [+0.026, +0.096] \\
\bottomrule\end{tabular}
\end{table}

We do not read this as an effect of pesticides. As described in Section~\ref{sec:data}, in the Kaggle table the pesticide column is an exact straight line in time for 13 of the 23 years, preceded by a peak around 2000. To a forest, such a column is a clock with a few kinks. The comparison with the calendar year supports this reading: a time variable alone recovers a gain of the same order. What the result does tell us is that the deviations from a linear trend are not random in time (maize yields in particular rise in a step between 2002 and 2005, Figure~\ref{fig:audit}c), and that a variable that carries timing information can pick some of it up. Mean temperature, the one genuine weather variable in the table, carries no information about those deviations. The longer FAOSTAT table supports this reading: the gain disappears there (Section~\ref{sec:faoresults}).

\subsection{Replication on the current FAOSTAT release}
\label{sec:faoresults}
The FAOSTAT table has 35 years and 70 observations, so the intervals are narrower than before, and it has no replicated rows. Table~\ref{tab:fao} gives the results, and the picture of Section~\ref{sec:honest} repeats. The per-crop linear trend has a leave-one-year-out RMSE of 0.292 Ton/Ha (interval 0.23--0.35) and a forward-chaining RMSE of 0.339 (0.25--0.43). Every model on feature set A$'$ lies between 0.84 and 1.06 under leave-one-year-out and between 0.93 and 1.15 under forward chaining, about three times worse. The crop mean scores 1.24 and 1.39. With the year added, the forest reaches 0.202 under leave-one-year-out (interval 0.15--0.25), XGBoost 0.24 and SVR 0.23, better than the trend when interpolating; Ridge, which cannot follow a non-linear trend, stays at 0.69. Under forward chaining the models with year (0.336, 0.318 and 0.366 for forest, XGBoost and SVR) tie with the trend (0.339).

\begin{table}[H]
\centering
\caption{Replication on the FAOSTAT table, 70 crop-year observations, 1990--2024. RMSE in Ton/Ha with bootstrap intervals over test years (35 for leave-one-year-out, 27 for forward chaining).}
\label{tab:fao}
\footnotesize
\begin{tabular}{lcccc}\toprule
 & \multicolumn{2}{c}{Leave-one-year-out} & \multicolumn{2}{c}{Forward chaining} \\\cmidrule(lr){2-3}\cmidrule(lr){4-5}
Model & RMSE [95\% CI] & R$^2$ & RMSE [95\% CI] & R$^2$ \\\midrule
\multicolumn{5}{l}{\emph{Baselines}} \\
Crop mean & 1.235 [1.05, 1.41] & 0.06 & 1.391 [1.15, 1.62] & -0.27 \\
Crop linear trend & 0.292 [0.23, 0.35] & 0.95 & 0.339 [0.25, 0.43] & 0.92 \\
\midrule\multicolumn{5}{l}{\emph{Feature set A$'$ (pesticide use, temperature change, crop)}} \\
Ridge & 0.979 [0.80, 1.15] & 0.41 & 1.117 [0.92, 1.32] & 0.18 \\
Random Forest & 0.909 [0.62, 1.19] & 0.49 & 0.992 [0.74, 1.25] & 0.35 \\
XGBoost & 1.061 [0.76, 1.34] & 0.30 & 1.152 [0.87, 1.43] & 0.13 \\
SVR & 0.843 [0.64, 1.03] & 0.56 & 0.931 [0.73, 1.13] & 0.43 \\
\midrule\multicolumn{5}{l}{\emph{Feature set C$'$ (A$'$ + year)}} \\
Ridge + year & 0.692 [0.58, 0.80] & 0.70 & 0.790 [0.64, 0.93] & 0.59 \\
Random Forest + year & 0.202 [0.15, 0.25] & 0.97 & 0.336 [0.25, 0.42] & 0.93 \\
XGBoost + year & 0.237 [0.17, 0.29] & 0.97 & 0.318 [0.22, 0.40] & 0.93 \\
SVR + year & 0.228 [0.17, 0.29] & 0.97 & 0.366 [0.28, 0.44] & 0.91 \\
\midrule\multicolumn{5}{l}{\emph{Trend plus residual model (features A$'$)}} \\
Trend + Ridge & 0.290 [0.22, 0.35] & 0.95 & 0.354 [0.27, 0.44] & 0.92 \\
Trend + Random Forest & 0.290 [0.22, 0.36] & 0.95 & 0.372 [0.28, 0.46] & 0.91 \\
Trend + XGBoost & 0.296 [0.23, 0.36] & 0.95 & 0.379 [0.28, 0.47] & 0.91 \\
Trend + SVR & 0.282 [0.22, 0.35] & 0.95 & 0.382 [0.28, 0.48] & 0.90 \\
\bottomrule\end{tabular}
\end{table}

The two-stage models, which learn what the trend misses, end differently here than on the Kaggle table (Table~\ref{tab:fao_ab}). Residual models built on pesticide use and temperature change give no improvement: their leave-one-year-out RMSE is 0.282--0.296 against 0.292 for the trend alone, and under forward chaining they are worse (0.354--0.382 against 0.339). In the ablation, temperature change has a gain of $-0.006$ (interval $-0.033$ to $0.022$) under leave-one-year-out and $-0.033$ ($-0.059$ to $-0.010$) under forward chaining; pesticide use gives $+0.021$ ($-0.014$ to $0.058$) and $-0.024$ ($-0.033$ to $-0.016$). The only residual feature that helps is the year itself ($+0.071$ and $+0.049$, both intervals above zero): the trend is not a straight line, as the step-like rise of maize in Figure~\ref{fig:fao} shows, and a time variable can describe that.

\begin{table}[H]
\centering
\caption{Which residual features help on the FAOSTAT table (Random Forest residual model). Gain is the reduction in RMSE relative to the linear trend, with a paired bootstrap 95\% interval over test years.}
\label{tab:fao_ab}
\footnotesize
\begin{tabular}{lcc}\toprule
Residual features (Random Forest) & Leave-one-year-out & Forward chaining \\
 & gain [95\% CI] & gain [95\% CI] \\\midrule
Crop indicator only & -0.000 [-0.001, -0.000] & +0.000 [-0.000, +0.001] \\
+ year & +0.071 [+0.044, +0.097] & +0.049 [+0.031, +0.068] \\
+ temperature change & -0.006 [-0.033, +0.022] & -0.033 [-0.059, -0.010] \\
+ pesticide use & +0.021 [-0.014, +0.058] & -0.024 [-0.033, -0.016] \\
+ temperature change + pesticide use & +0.002 [-0.029, +0.033] & -0.033 [-0.058, -0.013] \\
\bottomrule\end{tabular}
\end{table}

This is the result the Kaggle table could not give us. There, pesticide use appeared to explain 0.085 Ton/Ha of the trend's error (Section~\ref{sec:detrend}). On the longer table, where the pesticide series is nearly constant, the gain is gone. We take this as evidence that the earlier gain came from the shape of an interpolated series and not from pesticide use. National temperature change, a documented weather variable, adds nothing either. That is not proof that weather is irrelevant, since an annual national anomaly is a crude summary of growing-season conditions in one province (see Limitations).

\subsection{Independent check on district-level wheat}
\label{sec:district}
\begin{figure}[H]
\centering
\includegraphics[width=\textwidth]{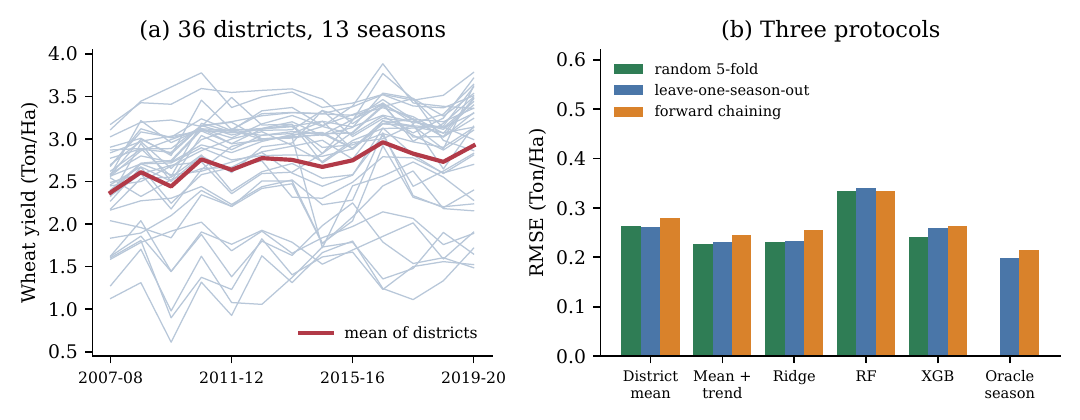}
\caption{District-level wheat panel, Punjab. (a) Yield of each district (thin lines) and the mean over districts (thick line). (b) RMSE of six predictors under three validation protocols; the oracle season shock uses test labels and is an upper bound.}
\label{fig:district}
\end{figure}

The district panel behaves very differently from the national table. Variation is mostly between places: district identity alone explains 85.4\% of the variance of yield, season alone 6.7\%, and the two together 91.5\%. Season-wide shocks are therefore small compared with the gap between, say, Okara and Chakwal (Figure~\ref{fig:district}a).

\begin{table}[H]
\centering
\caption{District-level wheat panel (464 observations). RMSE in Ton/Ha. The oracle uses the held-out season's own labels and is not a usable model. Bootstrap intervals are over test seasons.}
\label{tab:district}
\footnotesize
\begin{tabular}{lrrrr}\toprule
 & Random 5-fold & \multicolumn{2}{c}{Leave-one-season-out} & Forward chaining \\\cmidrule(lr){3-4}
Model & RMSE & RMSE [95\% CI] & R$^2$ & RMSE [95\% CI] \\\midrule
District mean & 0.263 & 0.261 [0.21, 0.31] & 0.83 & 0.279 [0.23, 0.33] \\
District mean + common linear trend & 0.228 & 0.231 [0.21, 0.25] & 0.87 & 0.246 [0.20, 0.29] \\
Ridge (district dummies + season index) & 0.232 & 0.234 [0.21, 0.26] & 0.86 & 0.255 [0.21, 0.29] \\
Random Forest (district + season index) & 0.334 & 0.340 [0.31, 0.37] & 0.71 & 0.334 [0.29, 0.38] \\
XGBoost (district + season index) & 0.240 & 0.258 [0.22, 0.29] & 0.83 & 0.264 [0.21, 0.32] \\
Oracle season shock (uses test labels) & -- & 0.199 [0.18, 0.22] & 0.90 & 0.216 [0.18, 0.25] \\
\bottomrule\end{tabular}
\end{table}

Table~\ref{tab:district} gives the results. Two things stand out. First, the inflation seen in the national table does not appear here: random 5-fold and leave-one-season-out give almost the same RMSE for every model (for example 0.263 and 0.261 for the district mean, and 0.334 and 0.340 for the forest). This is what one would expect when the rows are genuinely distinct observations and the season shock is small, and it supports our reading that the national table's inflation came from the nine-fold replication and not from temporal structure alone. Second, simple structure is hard to beat. The district mean with a common trend (RMSE 0.231 under leave-one-season-out) is as good as Ridge (0.234), and better than XGBoost (0.258) and the forest (0.340); under forward chaining the ordering is the same. The forest, which cannot extrapolate along the season index, is the worst.

The oracle row shows the room that remains. If the season-wide shock were known exactly, the RMSE of the district mean would fall from 0.261 to 0.199 Ton/Ha under leave-one-season-out (and from 0.279 to 0.216 under forward chaining). A model that could predict that shock from weather therefore has a ceiling of roughly a quarter of the error of the district mean, and the practical gain would be smaller. We could not test weather directly, because the file has no weather variables; this is the main reason we think district-level weather is the most useful data to add.

\subsection{Effect of the leaf score}
\label{sec:health}
\begin{figure}[H]
\centering
\begin{minipage}[c]{0.46\textwidth}
\includegraphics[width=\textwidth]{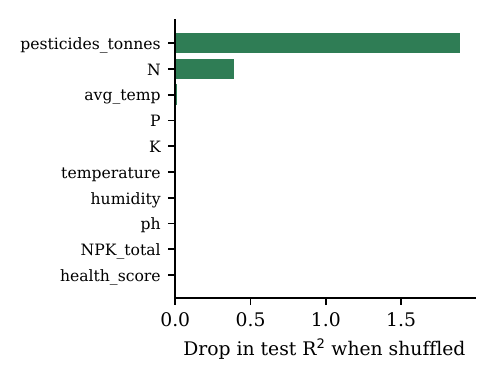}
\end{minipage}\hfill
\begin{minipage}[c]{0.50\textwidth}
\caption{Permutation importance of the ten features for the deployed XGBoost model on its 20\% test split. Only pesticide use, nitrogen and (slightly) temperature matter. Both leading features are constant within a crop-year or crop, so they behave as identifiers.}
\label{fig:perm}
\end{minipage}
\end{figure}

Table~\ref{tab:health} compares feature sets A and B under the year-aware protocols. The leaf score changes RMSE by at most 0.001 Ton/Ha. XGBoost and the ensemble give the same predictions with and without it, to three decimals. This is expected from the construction in Section~\ref{sec:integration}: the score is a function of the crop, and the nutrient features already encode the crop. A permutation test on the deployed XGBoost model agrees (Figure~\ref{fig:perm}): shuffling \code{health\_score} costs nothing, while shuffling pesticide use or nitrogen costs a great deal. Both of those are constant within a crop-year or within a crop, so they act as identifiers of the series rather than as agronomic drivers, and the pesticide column is, for 2001--2013, an exact line in time (Section~\ref{sec:audit}). We would not read the importance of pesticides as a causal effect.

\begin{table}[H]
\centering
\caption{Adding the leaf score (set B) to the nine tabular features (set A), crop-year table.}
\label{tab:health}
\small
\begin{tabular}{llrrrr}\toprule
Protocol & Model & RMSE (A) & RMSE (B) & R$^2$ (A) & R$^2$ (B) \\\midrule
Leave-one-year-out & Random Forest & 0.593 & 0.594 & 0.367 & 0.364 \\
Leave-one-year-out & XGBoost & 0.651 & 0.651 & 0.236 & 0.236 \\
Leave-one-year-out & Stacking ensemble & 0.737 & 0.737 & 0.023 & 0.021 \\
Forward chaining & Random Forest & 0.713 & 0.713 & -0.070 & -0.069 \\
Forward chaining & XGBoost & 0.817 & 0.817 & -0.405 & -0.405 \\
Forward chaining & Stacking ensemble & 0.956 & 0.958 & -0.923 & -0.928 \\
\bottomrule\end{tabular}
\end{table}

\paragraph{Sensitivity of the deployed application.}
We also queried the saved models directly. With default slider values (pesticides 6{,}000\,t, temperature 24\,$^\circ$C, N 56, P 42, K 61, soil temperature 24\,$^\circ$C, humidity 86\%, pH 6.4), we changed only the leaf score (Table~\ref{tab:sens}). Moving it across its entire range shifts the ensemble prediction by 0.004 Ton/Ha, XGBoost by nothing, and Random Forest by 0.043 Ton/Ha. In its present form, the leaf photograph is displayed and passed to the models, but it has almost no effect on the predicted yield.

\begin{table}[H]
\centering
\caption{Predicted yield (Ton/Ha) of the deployed models for different leaf scores, other inputs fixed.}
\label{tab:sens}
\small
\begin{tabular}{lcccc}
\toprule
Leaf input (\code{health\_score}) & RF & XGBoost & SVR & Ensemble \\
\midrule
Healthy leaf (0.9921) & 1.660 & 1.854 & 2.352 & 1.824 \\
No image (0.5) & 1.660 & 1.854 & 2.343 & 1.825 \\
Diseased leaf (0.0007) & 1.703 & 1.854 & 2.326 & 1.828 \\
\bottomrule
\end{tabular}
\end{table}

\subsection{Leaf-health classifier}
\label{sec:cvresults}
\begin{figure}[H]
\centering
\begin{subfigure}[t]{0.36\textwidth}
\includegraphics[width=\textwidth]{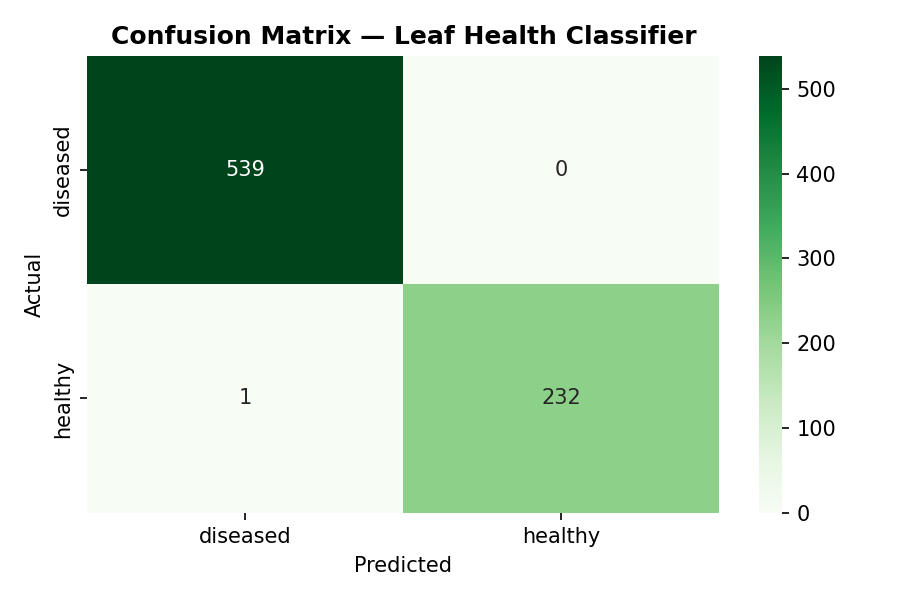}
\caption{Confusion matrix (validation set)}
\end{subfigure}\hfill
\begin{subfigure}[t]{0.60\textwidth}
\includegraphics[width=\textwidth]{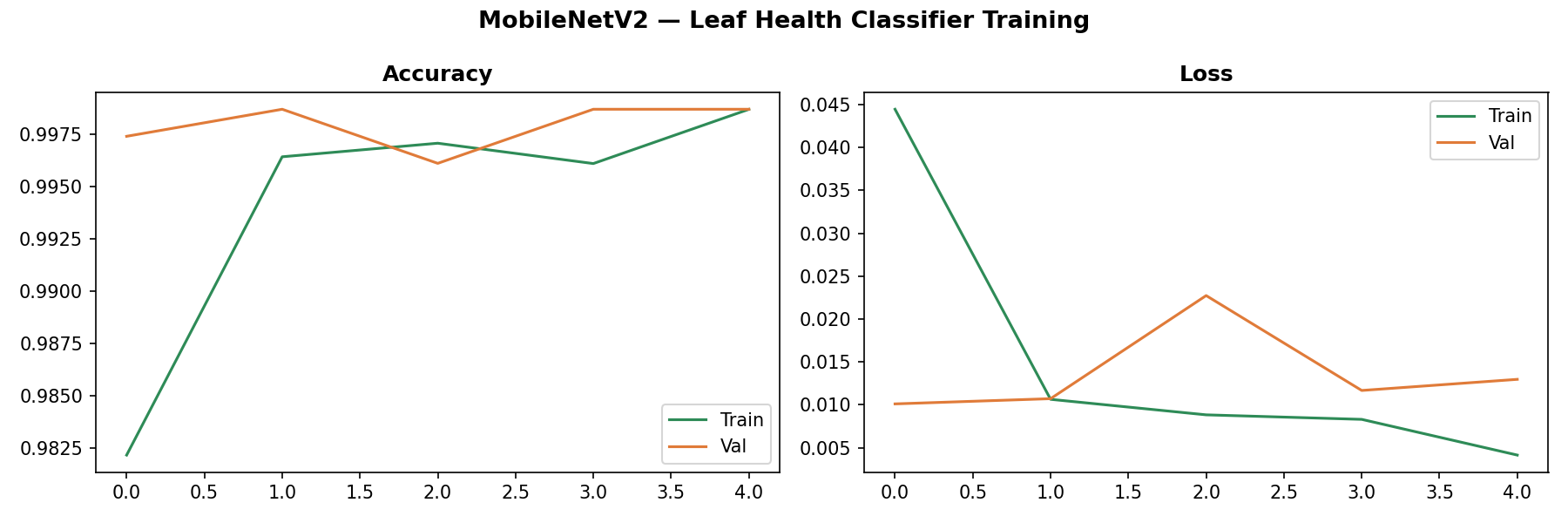}
\caption{Accuracy and loss over five epochs}
\end{subfigure}
\caption{Leaf-health classifier: evaluation on 772 held-out images (539 diseased, 233 healthy) and training history.}
\label{fig:cm}
\end{figure}

On the 772 validation images the classifier made a single error: one healthy leaf was labelled diseased (Figure~\ref{fig:cm}). All 539 diseased leaves and 232 of 233 healthy leaves were classified correctly, an accuracy of $771/772 = 99.87\%$ with a Wilson 95\% interval of 99.27--99.98\%. Table~\ref{tab:cv} gives per-class precision, recall and F1, computed from the confusion matrix. Training converged quickly (Figure~\ref{fig:cm}b): validation accuracy was about 99.7\% after the first epoch and reached 99.87\% in the second. Training accuracy starts lower than validation accuracy, which is what one expects when training images are augmented and dropout is active.

\begin{table}[H]
\centering
\caption{Per-class results of the leaf classifier on the validation set.}
\label{tab:cv}
\small
\begin{tabular}{lcccc}
\toprule
Class & Precision & Recall & F1 & Support \\
\midrule
Diseased & 0.9981 & 1.0000 & 0.9991 & 539 \\
Healthy & 1.0000 & 0.9957 & 0.9979 & 233 \\
Macro average & 0.9991 & 0.9979 & 0.9985 & 772 \\
\bottomrule
\end{tabular}
\end{table}

Three caveats limit what this number means. First, PlantVillage images are photographed against plain backgrounds in controlled light, and the collection contains several pictures of the same leaves, so a random image-level split can put near-duplicates on both sides. Second, the same validation set was used to choose the stopping epoch and to report accuracy, which is known to bias estimates upward \citep{cawley2010}; we held out no separate test set. Third, the validation images are laboratory photographs, and accuracy has been reported to drop markedly when models trained on such images meet other sources \citep{mohanty2016}; we address this below with images from other sources, but not with photographs taken in Bahawalpur. We therefore read 99.87\% as an optimistic estimate for clean maize images. As a quick sanity check on a non-leaf input, the deployed classifier labelled an image of random noise ``healthy'' with a probability of 0.535, so its output should not be trusted outside the domain it was trained on.

\paragraph{Unseen images.} Table~\ref{tab:field} shows the results on the 336 images that remained after removing near-duplicates of PlantVillage (Section~\ref{sec:unseenmethods}). The classifier recognised 324 of the 336 diseased leaves, a recall of 96.4\% (Wilson 95\% interval 93.9--97.9\%). Recall was highest for northern leaf blight (98.1\%) and lowest for common rust (93.9\%). The 12 missed leaves received healthy probabilities between 0.59 and 0.92. On the contamination control, the 800 near-duplicates of training images were classified perfectly; that is the number one would have reported by testing on the whole Kaggle dataset without removing duplicates. Two limits remain: the unseen set contains no healthy leaves, so we know the sensitivity of the classifier on another image source but not its false-alarm rate, and no image comes from Bahawalpur.

\begin{table}[H]
\centering
\caption{Recall of the diseased class on images that are not in PlantVillage, and on near-duplicates of PlantVillage training images (Corn or Maize Leaf Disease Dataset).}
\label{tab:field}
\footnotesize
\begin{tabular}{p{6.2cm}cc}\toprule
Image set & Correct / total & Recall of diseased [95\% CI] \\\midrule
Unseen: northern leaf blight & 158 / 161 & 0.981 \\
Unseen: common rust & 107 / 114 & 0.939 \\
Unseen: gray leaf spot & 59 / 61 & 0.967 \\
\textbf{Unseen: all diseased} & \textbf{324 / 336} & \textbf{0.964 [0.939, 0.979]} \\\midrule
Near-duplicates of training images (contaminated; sample of 800) & 800 / 800 & 1.000 [0.995, 1.000] \\
\bottomrule\end{tabular}

\end{table}

\clearpage
\section{Findings}
\label{sec:findings}
Table~\ref{tab:findings} collects what we found, in the order in which the evidence appears. Each row gives the finding, the numbers behind it and where to look.

\begin{table}[H]
\centering
\caption{Summary of findings. RMSE in Ton/Ha; LOYO is leave-one-year-out; ``ensemble'' is the Ridge-stacked model.}
\label{tab:findings}
\footnotesize
\setlength{\tabcolsep}{4pt}
\begin{tabular}{@{}c p{6.1cm} p{6.3cm} c@{}}
\toprule
 & Finding & Key evidence & See \\
\midrule
1 & The 414-row national table holds only 46 independent observations. & $2$ crops $\times$ 23 years $\times$ 9 copies; each copy has the same yield. & \ref{sec:audit} \\
2 & Random splits inflate the scores. & XGBoost $R^2$ 0.994 (random 5-fold) $\to$ $-0.204$ (LOYO) on the same rows; 0.54 on the deduplicated table. & \ref{sec:leak}, \ref{sec:honest} \\
3 & A per-crop linear trend beats every model that is not given the year. & LOYO RMSE 0.243 against 0.59--0.74 (46 rows); 0.292 against 0.84--1.06 on the FAOSTAT table (70 rows). & \ref{sec:honest}, \ref{sec:faoresults} \\
4 & Giving the models the year helps interpolation but not forecasting. & Random Forest + year: LOYO 0.141 and 0.202 (FAOSTAT), but forward chaining 0.333 and 0.336 against 0.317 and 0.339 for the trend (ties). & \ref{sec:honest}, \ref{sec:faoresults} \\
5 & The pesticide series is imputed in both data releases, and the gain it seemed to give on the Kaggle table does not survive on the longer FAOSTAT table; temperature change adds nothing. & Kaggle: exact line, $-989.2$ t per year, gain 0.085. FAOSTAT: 4 distinct values, 27 of 35 imputed; gain $+0.021$ [$-0.014$, $0.058$] (LOYO), $-0.024$ (forward). & \ref{sec:audit}, \ref{sec:detrend}, \ref{sec:faoresults} \\
6 & In an independent district-level wheat panel, most variation is spatial and simple structure is hard to beat. & District identity explains 85.4\% of variance, season 6.7\%; district mean with common trend 0.231; known season shock would give 0.199. & \ref{sec:district} \\
7 & The leaf classifier is accurate on its benchmark and transfers reasonably to unseen diseased leaves, but the yield models do not use it in any meaningful way. & 99.87\% on PlantVillage validation; 324/336 (96.4\%) recall on unseen images; score changes RMSE by $\leq 0.001$; ensemble moves 0.004 Ton/Ha. & \ref{sec:health}, \ref{sec:cvresults} \\
8 & The stacking ensemble is effectively XGBoost and is not better than its parts under honest validation. & Meta-learner weight 0.967 on XGBoost; LOYO RMSE 0.737 against 0.59--0.65 for the single models. & \ref{sec:honest} \\
\bottomrule
\end{tabular}
\end{table}

\paragraph{What we do not claim.}
We do not claim that weather has no effect on yield: in our tables rainfall is constant, soil values are crop-level averages and temperature is a national mean of unclear origin, so a null result is what one would expect whatever the truth. We do not claim that learned models cannot beat a trend in general, only that they did not do so here without the year. We do not claim that leaf images cannot help to predict yield, only that our data cannot test it. We do not know how often the classifier raises false alarms on healthy field leaves, because our unseen images are all diseased. And we do not read the importance of pesticide use as an agronomic effect.

\section{Discussion}
Three lessons come out of this work, and we think they apply beyond our table.

\paragraph{Check what a row is.}
The most valuable step in this project was also the simplest: counting how many distinct targets the table contains. Our nine-fold replication came from a many-to-one join that is easy to overlook, because the result still looks like a clean rectangular table with no missing values. A group-aware split, or even a simple check that no two rows share a target and a year, would have caught it at the start. When the unit of prediction is a crop-year, the split should be done by crop-year or, as here, by year.

\paragraph{Always include a trend baseline.}
Yield series rise over time, so a model that is not told the year is competing against a baseline it cannot see. That simple methods are hard to beat is a recurring finding in forecasting \citep{makridakis2018}, and it held again on the independent district panel, where a district mean with a common trend matched or beat every learned model. In our data, a straight line per crop beats nine-feature machine-learning models by a wide margin. Reporting a model without such a baseline gives no sense of whether the weather and soil variables add anything. In our case, on this table, they do not add much, and the one that seemed to add something (pesticide use) turned out to be an imputed series whose apparent gain vanished on a longer table; in fairness the variables available are weak: rainfall is constant, soil values are crop averages, temperature is a national mean with an undocumented spread, and pesticide use is one number per year.

\paragraph{Be careful what an image feature can claim.}
The leaf classifier is accurate on its own task, and a health label is useful information for a farmer in its own right, and on diseased leaves it never saw in training it still recovered 96\% (Section~\ref{sec:cvresults}); we cannot yet say how often it would raise a false alarm on healthy field leaves. But linking it to yield requires data we did not have: leaf images and yields from the same fields and seasons. Without them, any feature we feed to the yield model is a stand-in, and our stand-in carries no new information. The correct conclusion is not that images cannot help yield prediction. It is that our data cannot tell, and that the interface in its present form should be presented as a leaf-health tool with a yield estimator beside it, not as a combined model.

\section{Hyperparameters and Reproducibility}
\label{sec:hyper}
Table~\ref{tab:hyper} lists every setting used in this paper. All random seeds are fixed at 42, hyperparameters were chosen once and not tuned, and the same settings are used under every validation protocol. The scripts that produce each table and figure are in the \texttt{code/} folder of the repository (Data and Code Availability).

\begin{table}[H]
\centering
\caption{Settings used for the models in this paper.}
\label{tab:hyper}
\small
\begin{tabular}{p{2.6cm}p{10.2cm}}
\toprule
Model & Settings \\
\midrule
Random Forest & 200 trees, max depth 8 (10 in the 13-feature baseline), seed 42 \\
XGBoost & 200 trees, learning rate 0.05, max depth 5 (6 in the baseline), seed 42 \\
SVR & RBF kernel, $C=100$, $\gamma=0.1$, $\epsilon=0.1$, standardised inputs \\
Ridge & $\lambda = 1$, standardised inputs \\
Stacking & RF, XGBoost, SVR $\to$ Ridge ($\lambda=1$), 5-fold out-of-fold predictions \\
Residual models (Section~\ref{sec:detrend}) & Ridge ($\lambda=1$); Random Forest (200 trees, depth 3, min.\ leaf 3); XGBoost (100 trees, depth 2, learning rate 0.05); SVR (RBF, $C=1$, $\epsilon=0.05$). Chosen once, not tuned. \\
Leaf classifier & MobileNetV2 (frozen), GAP, Dense 128 ReLU, Dropout 0.3, Dense 1 sigmoid; Adam, binary cross-entropy; early stopping (patience 3) on validation accuracy \\
Evaluation & Random 5-fold (5 repeats); leave-one-year-out (23 folds); forward chaining (test years 1998--2013); bootstrap over years, 2{,}000 resamples; on the FAOSTAT table, 35 leave-one-year-out folds and forward-chaining test years 1998--2024 \\
\bottomrule
\end{tabular}
\end{table}

\section{Limitations}
\begin{itemize}
  \item \textbf{Sample size.} The Kaggle-derived table has 46 independent observations from 23 years and the FAOSTAT table 70 from 35 years, in both cases for two crops. Confidence intervals are wide and model rankings among the learned models are not reliable.
  \item \textbf{Aggregation level.} Yields, pesticides and temperature are national or crop-level figures. Soil and humidity are crop-level means and carry no spatial information about Punjab's districts, so the models cannot represent regional variation.
  \item \textbf{National weather.} The only weather variable we could test, annual national temperature change, cannot describe growing-season conditions in one province, so its null result is weak evidence about weather in general.
  \item \textbf{Period.} The Kaggle-derived table ends in 2013. The FAOSTAT table reaches 2024 but is national and has only 70 observations. The district panel reaches 2019--20 but covers wheat only, has no weather variables, and we did not test generalisation to unseen districts.
  \item \textbf{Pesticide series.} The Kaggle table's pesticide series is an exact straight line from 2001, and the current FAOSTAT series is mostly imputed (27 of 35 values) and takes only four distinct values. Neither is usable as a measure of annual pesticide use, and any apparent link to yield should be regarded as an artefact.
  \item \textbf{Temperature source.} The meaning of the nine yearly readings is not documented in the files we used. We averaged them, which may or may not be the right summary.
  \item \textbf{Constructed score.} \code{health\_score} in the training data is a rescaled \code{NPK\_total}. The integration therefore adds nothing, and the deployed application's response to a leaf image is nearly flat.
  \item \textbf{Classifier evaluation.} Maize only. The PlantVillage validation set was also used to select the stopping epoch. The unseen-image test contains diseased leaves only, its provenance (presumably PlantDoc) is inferred from the dataset description and the image counts, and no photographs from Bahawalpur were tested, so the false-alarm rate on healthy field leaves is unknown.
  \item \textbf{Tuning.} Hyperparameters were fixed once and not tuned; a tuned SVR might close part of its gap, although tuning on 46 rows would itself require nested validation.
  \item \textbf{Single seed.} We report one seed for the classifier and the deployed models; the random 5-fold results average five repeats.
  \item \textbf{Not for decisions.} The system is an academic prototype and has not been validated for farming, financial or policy decisions.
\end{itemize}

\section{Conclusion and Future Work}
We built a yield-prediction prototype for Punjab that combines tree ensembles, SVR and a MobileNetV2 leaf-health classifier, and we were able to reproduce very high scores with it, up to $R^2=0.991$. We then found that these scores came from nine-fold replication in the merged table and did not reflect real predictive skill. Evaluated by year, a linear trend per crop is the strongest option available without extra inputs (RMSE 0.24 Ton/Ha leave-one-year-out on the Kaggle-derived table and 0.29 on a longer table rebuilt from FAOSTAT); models given the year match or beat it, and models without the year do not. Neither pesticide use, whose series is imputed in both releases, nor national temperature change improves on the trend. An independent district-level wheat panel confirmed that most variation in Punjab is spatial and that simple structure is hard to beat, and it bounded what a perfect season-wide predictor could add. The leaf classifier works well on its own benchmark (99.87\%) and recovered 96.4\% of 336 diseased leaves it had never seen, but our attempt to feed it into the yield models failed for lack of paired data.

The path forward is fairly clear. We would (i) extend the district-by-season yield panel to more crops and add district-level seasonal weather and vegetation indices, which would test the weather question that national data cannot answer; (ii) collect or find leaf images with matched yield records so that \code{health\_score} can be measured and its value tested; (iii) report leave-one-year-out and forward-chaining results as the headline metrics, with a trend baseline alongside; and (iv) test the classifier on healthy and diseased photographs from the Bahawalpur region, and on wheat. The code, the crop-year tables, the Colab notebook that rebuilds the FAOSTAT table and runs the unseen-image test, and the deployed application are available so that others can reproduce what we did and, where we were wrong, correct it.

\section*{Data and Code Availability}
The code, notebook, trained models and result tables are available at \url{https://github.com/AminahAsif/Crop_Yield_Prediction-_Punjab} 
and the demo at \url{https://huggingface.co/spaces/AminahAsif/Crop-yield-punjab}. The deduplicated crop-year table, the extended FAOSTAT table, the cleaning code for the district panel and a Colab notebook that rebuilds the FAOSTAT table and runs the unseen-image test are included in the repository. The source datasets are public on Kaggle (Table~\ref{tab:data}); please check their licences before redistributing the raw files. The district yield file is available from the Open Data Pakistan portal \citep{pbsdistrict}, which lists no licence, so we do not redistribute it.

\section*{Acknowledgements}
During the preparation of this work, the authors used Claude (Anthropic) to assist with code development and debugging.
The authors reviewed and edited all content as needed and take full responsibility for the content of this paper.

\setlength{\bibsep}{2pt plus 0.3ex}
\small
\bibliographystyle{plainnat}
\bibliography{references}

\end{document}